\documentclass{article}

\usepackage[final,nonatbib]{neurips_2022}
\usepackage[numbers]{natbib}

\usepackage[T1]{fontenc}    
\usepackage{hyperref}       
\usepackage{url}            
\usepackage{booktabs}       
\usepackage{amsfonts}       
\usepackage{nicefrac}       
\usepackage{microtype}      
\usepackage{xcolor}         
\usepackage[nolist,nohyperlinks]{acronym}
\usepackage{amsmath}

\usepackage{graphicx}
\usepackage{caption}
\usepackage{subcaption}
\usepackage{wrapfig}

\usepackage{siunitx}

\title{\mbox{How Crucial is Transformer in Decision Transformer?}}

\author{%
  Max Siebenborn \\
  Technical University of Darmstadt \\
  Department of Computer Science \\
  \texttt{max.siebenborn@stud.tu-darmstadt.de} \\
  \And
   Boris Belousov \\
   Technical University of Darmstadt \\
   German Research Center for AI (DFKI)\\
   Systems AI for Robot Learning Group \\
   \And
   Junning Huang \\
   Technical University of Darmstadt \\
   Department of Computer Science \\
   Intelligent Autonomous Systems Group \\
   \And
   Jan Peters \\
   Technical University of Darmstadt \\
   German Research Center for AI (DFKI) \\
  Hessian.AI \& Centre for Cognitive Science \\
}

\begin{document}

\maketitle
\vspace{-1em}

\begin{abstract}
  Decision Transformer (DT) is a recently proposed architecture for Reinforcement Learning that frames the decision-making process as an auto-regressive sequence modeling problem and uses a Transformer model to predict the next action in a sequence of states, actions, and rewards.
  In this paper, we analyze how crucial the Transformer model is in the complete DT architecture on continuous control tasks.
  Namely, we replace the Transformer by an LSTM model while keeping the other parts unchanged to obtain what we call a \emph{Decision LSTM} model. 
  We compare it to DT on continuous control tasks, including pendulum swing-up and stabilization, in simulation and on physical hardware.
  Our experiments show that DT struggles with continuous control problems, such as inverted pendulum and Furuta pendulum stabilization.
  On the other hand, the proposed Decision LSTM is able to achieve expert-level performance on these tasks, in addition to learning a swing-up controller on the real system.
  These results suggest that the strength of the Decision Transformer for continuous control tasks may lie in the overall sequential modeling architecture and not in the Transformer per se.
\end{abstract}

\section{Introduction}

Transformers \cite{vaswani2017attention} have shown impressive results across a number of problem domains in Natural Language Processing~\cite{devlin2019bert, Radford2018ImprovingLU, brown-transformer-nlp} and Computer Vision \cite{vision-transformer, Khan_2022}.
Inspired by these results, \cite{chen2021decision, janner2021offline} framed Reinforcement Learning (RL) as a sequence modeling problem, in which Transformer predicts the next element in a sequence of states, actions and rewards.
In \cite{chen2021decision}, the Decision Tranformer (DT) is proposed, an offline RL algorithm that auto-regressively models trajectories using the GPT-2 architecture~\cite{Radford2019LanguageMA}.
The Trajectory Transformer architecture from~\cite{janner2021offline} is similar to DT but instead of return-to-go values it utilizes beam-search planning for sequence generation and employs state and reward prediction as well as discretization.
The evaluations in~\cite{chen2021decision} showed that DT is stronger than straightforward Behavior Cloning (BC) on the D4RL dataset~\cite{fu2021d4rl}, which includes discrete Atari games and continuous control tasks from OpenAI gym~\cite{brockman2016openai}.
However, from these experiments it remains unclear whether Decision Transformer is also competitive for dynamic tasks that require stabilization of systems around an unstable equilibrium, as well as for real robot control tasks.

In this paper, we evaluate Decision Transformer on robot learning tasks, focusing on two aspects.
First, we evaluate DT on \emph{stabilization tasks}---on various pendulum swing-up and stabilization environments.
The goal of an agent in these tasks is to reach an unstable equilibrium and stabilize the system around it.
Second, we validate our simulation results on a \emph{real robotic platform}.
This evaluation is crucial since the gap between simulation and reality is still an open issue in robotics and RL~\cite{muratore2022robot,zhao-simtoreal}, and the results in simulation do not directly transfer to reality.
Furthermore, for real robotic applications, the model inference time must be sufficiently small to enable real-time control.

In addition to our evaluation of Decision Transformer, we propose a related architecture, which we call Decision LSTM (DLSTM), that builds on top of DT but replaces the GPT-2 Transformer by an LSTM \cite{lstms-schmidhuber} model.
We use DLSTM to test whether framing RL as a sequence modeling problem allows using other architectures apart from Transformers, and whether they can yield better results.
We compare the performances of both DT and DLSTM against a straightforward Behavior Cloning (BC) model which mimics actions based on the observed states without taking rewards into account.

In summary, our paper provides the following three contributions.
\begin{itemize}
    \item The introduction of Decision LSTM as a novel architecture for offline RL, which builds on top of Decision Transformer. DLSTM shows the general capabilities of framing RL as a sequence modeling problem independent of the concrete architecture.
    \item An evaluation of DT and DLSTM in continuous control tasks that require fine-grained stabilization.
    Experiments on a Furuta pendulum platform highlight the issues and trade-offs of the different architectures when deployed in the real world.
    \item A thorough investigation concentrated on whether the functionalities and effects of critical ingredients of the Decision Transformer, such as the return-to-go values, can be validated in the continuous control environments.
\end{itemize}

\section{Background} \label{sec:background}

\begin{figure}[t]
     \centering
     \includegraphics[width=\textwidth]{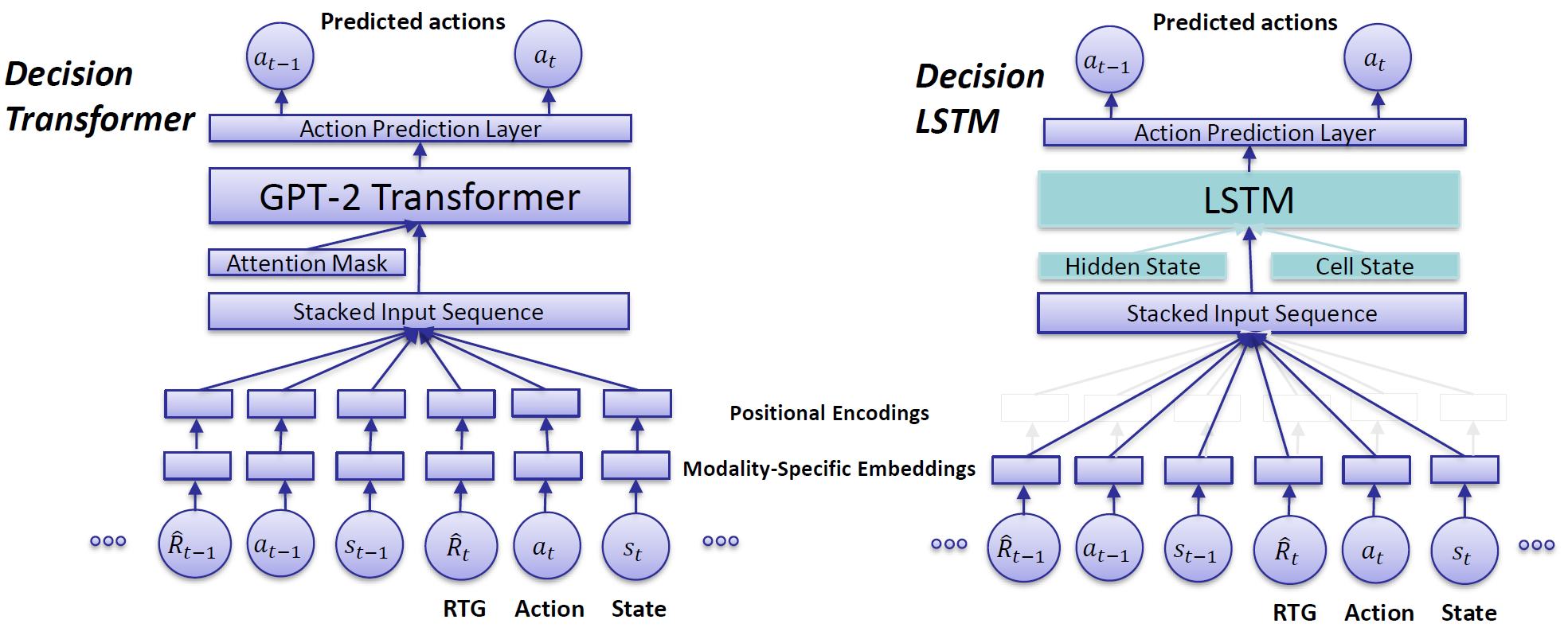}
     \caption{Comparison of the architectures of Decision Transformer (on the left) and the proposed Decision LSTM (on the right). The bottom items show the sequences of observed return-to-go (RTG) values, actions and states, which are used to predict the next action in an auto-regressive way. The key differences between the architectures are the use of an LSTM network as a replacement for the GPT-2 Transformer and the removal of positional encodings for DLSTM.}
     \label{fig:arch_comp}
     \vspace{-1em}
\end{figure}

A key motivation behind the Decision Transformer~\cite{chen2021decision} architecture is to frame reinforcement learning as a sequence modeling problem~\cite{Bai2018AnEE}.
Sequence modeling is predominant in natural language processing, where, e.g., a sentence can be seen as a sequence of words.
Language models predict the next word in a sentence by taking the previous words as input \cite{sutskever2014sequence}.
Similarly, DT predicts the next action in a sequence of states, actions, and rewards.
Recurrent Neural Networks (RNNs) \cite{rumelhart:errorpropnonote, osti_6910294} and especially LSTMs \cite{lstms-schmidhuber} have been considered state-of-the-art sequence models thanks to their ability to process sequences of varying length and make information from previous timesteps persistent inside the network.
However, such sequential processing of data precludes parallelization of RNN and LSTM computations, resulting in potentially long training times \cite{vaswani2017attention}.

Recently Transformers \cite{vaswani2017attention} have become predominant in natural language processing and sequence modeling. 
Transformer is an architecture for auto-regressive sequence modeling that is purely based on the \textit{attention} mechanism and does not contain any recurrent or convolutional structures.
In contrast to RNNs and LSTMs, they are highly parallelizable since their attention mechanism does not require sequential processing of the input elements.
Furthermore, models such as BERT \cite{devlin2019bert}, and the GPT-x architectures \cite{Radford2018ImprovingLU, Radford2019LanguageMA} have shown the capabilities of Transformers to build large pre-trained models that can be finetuned on specific tasks.

In terms of computational complexity, Transformer layers are linear in the input dimensionality and quadratic in the input length \cite{vaswani2017attention}.
This is advantageous for short input sequences with high-dimensional latent representations (common in NLP) but can be problematic for long input sequences (e.g., sequences of states, actions, and rewards in RL problems).
Recurrent networks, however, can have computational advantages for long input sequences with small input dimensionality because they scale linearly in the input length and quadratically in the input dimensionality.

Decision Transformer~\cite{chen2021decision} brings the benefits of sequence modeling to model-free offline reinforcement learning.
It frames RL as a prediction problem, with the goal of predicting the next action given the history of past transitions in a trajectory
\begin{equation*} \label{eq:dt-traj-repr}
\tau = (\hat{R}_{1}, s_{1}, a_{1}, \hat{R}_{2}, s_{2}, a_{2}, ... , \hat{R}_{T}, s_{T}, a_{T}) \,.
\end{equation*}
Here, $\hat{R}_{t}$ is the return-to-go (RTG) value, i.e., the sum of the future rewards in the trajectory,  $s_{t}$ is the state, and $a_{t}$ is the action at time step $t$, respectively.
DT aims to solve the RL problem without making use of conventional value-based methods such as Dynamic Programming or TD-Learning~\cite{chen2021decision}
by conditioning outputs on RTG values, similar to related return-conditioned approaches~\cite{liu2022goalconditioned, kumar2019rewardconditioned, schmidhuber2020reinforcement}.

\section{Methods} \label{sec:experiment-methodology}

\begin{figure}[t]
  \centering
  \begin{tabular}[c]{cccc}
    \begin{subfigure}[b]{0.22\textwidth}
      \includegraphics[width=\textwidth]{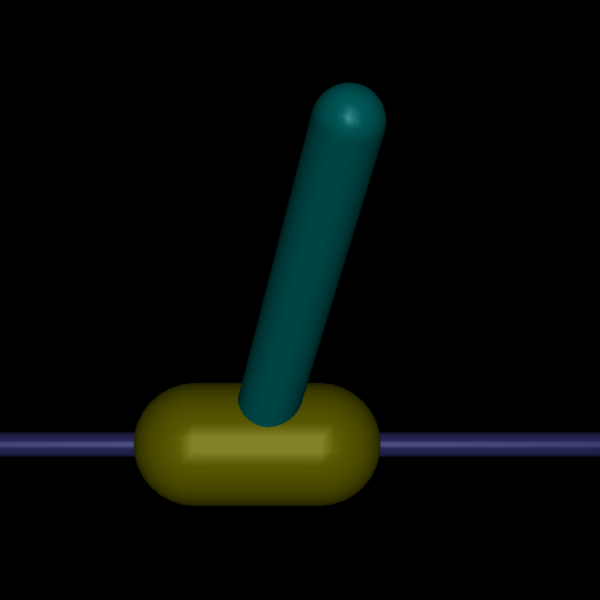}
      \caption{Mujoco Inverted Pendulum (Stabilization).}
      \label{fig:env_mujoco_pendulum}
    \end{subfigure}&
    \begin{subfigure}[b]{0.22\textwidth}
      \includegraphics[width=\textwidth]{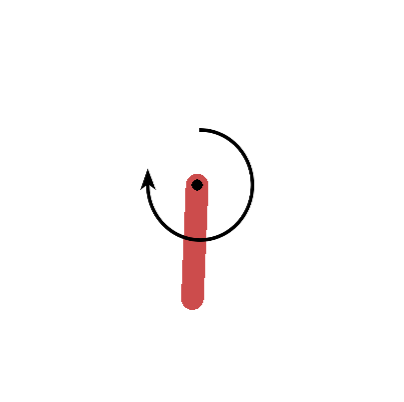}
      \caption{OpenAI Pendulum.}
      \label{fig:env_openai_pendulum}
    \end{subfigure}&
    \begin{subfigure}[b]{0.22\textwidth}
      \includegraphics[width=\textwidth]{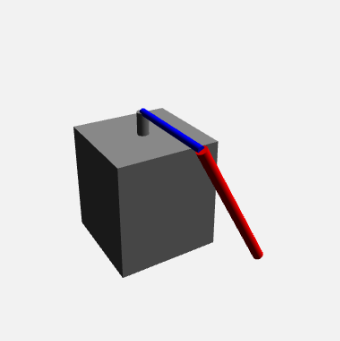}
      \caption{Furuta Pendulum (Simulation).}
      \label{fig:env_furuta_sim}
    \end{subfigure}&
    \begin{subfigure}[b]{0.22\textwidth}
      \includegraphics[width=\textwidth]{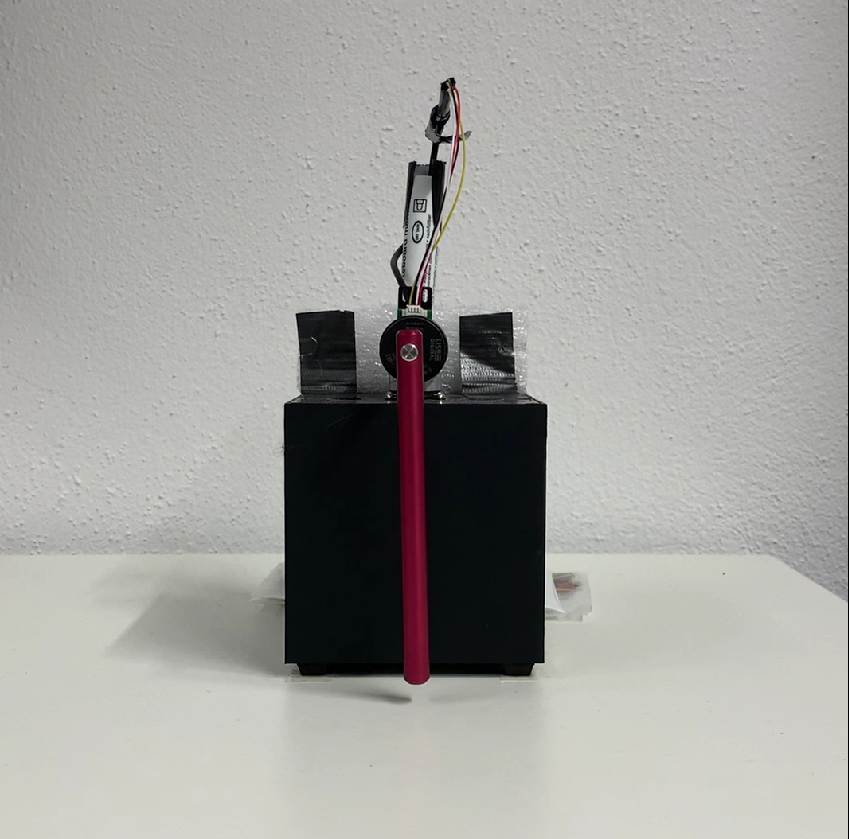}
      \caption{Furuta Pendulum (Real Hardware).}
      \label{fig:env_furuta_rr}
    \end{subfigure}
  \end{tabular}
  \caption{Experiment environments to test the capabilities of Decision Transformer and Decision LSTM on continuous control tasks requiring stabilization.}\label{fig:envs}
  \vspace{-1em}
\end{figure}

The main question addressed in this paper is whether Transformer is crucial for the Decision Transformer architecture.
We hypothesize that the overall framing of RL as a sequence modeling problem is more responsible for the strong performance of DT than the use of the Transformer.
To prove this point, we introduce Decision LSTM (DLSTM), a novel architecture which builds on the Decision Transformer but replaces the GPT-2 model by an LSTM network.
The DT and DLSTM architectures are shown in Figure~\ref{fig:arch_comp}.
DLSTM introduces the following architectural adjustments:
\begin{itemize}
    \item the GPT-2 Transformer is replaced by an LSTM;
    \item the attention mask is removed because LSTM does not utilize it;
    \item positional embeddings are removed because LSTM processes inputs sequentially;
    \item LSTM's hidden states and cell states are initialized with zero vectors.
\end{itemize}
The LSTM architecture has proven to be a successful tool for sequence modeling~\cite{staudemeyer-lstm}.
It provides computational benefits especially on long sequences, because it scales linearly with the input length whereas Transformer scales quadratically.
Therefore, gains in performance and real-time capability are expected from DLSTM in RL, where input lengths may range between $100$'s to $1000$'s timesteps.

We evaluate DT and DLSTM on several continuous control tasks, which are shown in Figure~\ref{fig:envs}.
Additionally, we report the results of a straightforward Behavior Cloning (BC) baseline which predicts the next action using a feedforward neural network trained on the dataset of past trajectories.
Our experimental methodology consists of the following steps.
First, a dataset of trajectories is collected using a behavior policy.
All three approaches---DT, DLSTM, and BC---operate in the offline mode.
Following the D4RL~\cite{fu2021d4rl} protocol, we collect separate datasets of \emph{expert} quality (behavior policy solves the task at expert level) and of \emph{replay} quality
(data from early epochs of training of an online model-free RL algorithm is mixed with data from late epochs).
On the replay data, DT and DLSTM are expected to perform better than BC, because they weigh experiences by the reward, whereas BC does not take the reward into account.
Second, all models are trained until convergence on the collected data.
Third, fully optimized models are evaluated over $30$ runs in the respective environments.
Our implementation is based on the original DT codebase with default parameters.

\section{Experiments}

In this section, we present and discuss the results of the experiments described in Section~\ref{sec:experiment-methodology}.

\paragraph{Simulation results on expert data.}

\begin{table}[t]
    \centering
    \caption[Simulation results on expert data.]{Simulation results on \emph{expert} data. Fully trained DT, DLSTM, and BC models are evaluated in 4 simulated environments. Mean episode return over the dataset $\overline{G_{\mathrm{Data}}}$ and the mean $\pm$ standard deviation of episode returns over $30$ evaluation episodes are reported.
    DLSTM outperforms DT in all cases, and it outperforms BC in 3/4 environments, performing on par in OpenAI pendulum swing-up.}
    {\fontsize{7}{7} \selectfont            
    \begin{tabular}{ccc|ccc} \toprule
    \multicolumn{3}{c}{} & \multicolumn{3}{c}{\textbf{Evaluation Episode Returns}}  \\
    \textbf{Environment} & \textbf{Dataset} & \textbf{$\overline{G_{\mathrm{Data}}}$} &  \textbf{DT} &  \textbf{DLSTM} &  \textbf{BC} \\ \hline 
    Mujoco Pendulum Stabilization & Expert & 1000.00 $\pm$ 0.00 & 454.72 $\pm$ 360.12 & \textbf{985.31 $\pm$ 71.96} & 61.61 $\pm$ 170.16 \\ 
    OpenAI Pendulum Swing-up & Expert & -207.53 $\pm$ 167.75 & -761.44 $\pm$ 375.71 & \textbf{-252.86 $\pm$ 233.21} & \textbf{-235.78 $\pm$ 204.45} \\ 
    Furuta Pendulum Stabilization & Expert & 5.95 $\pm$ 0.02 & 0.46 $\pm$ 0.03 & \textbf{5.93 $\pm$ 0.01} & 1.82 $\pm$ 1.60 \\ 
    Furuta Pendulum Swing-up & Expert & 2.93 $\pm$ 0.63 & 0.74 $\pm$ 0.24 & \textbf{1.79 $\pm$ 1.12} & 0.87 $\pm$ 0.21 \\ \bottomrule
    \end{tabular}
    \label{tab:experiment-final-results}
    }
    \vspace{-1em}
\end{table} 

\begin{figure}[t]
    \centering
    \begin{subfigure}[b]{0.475\textwidth}
        \centering
        \includegraphics[width=\textwidth]{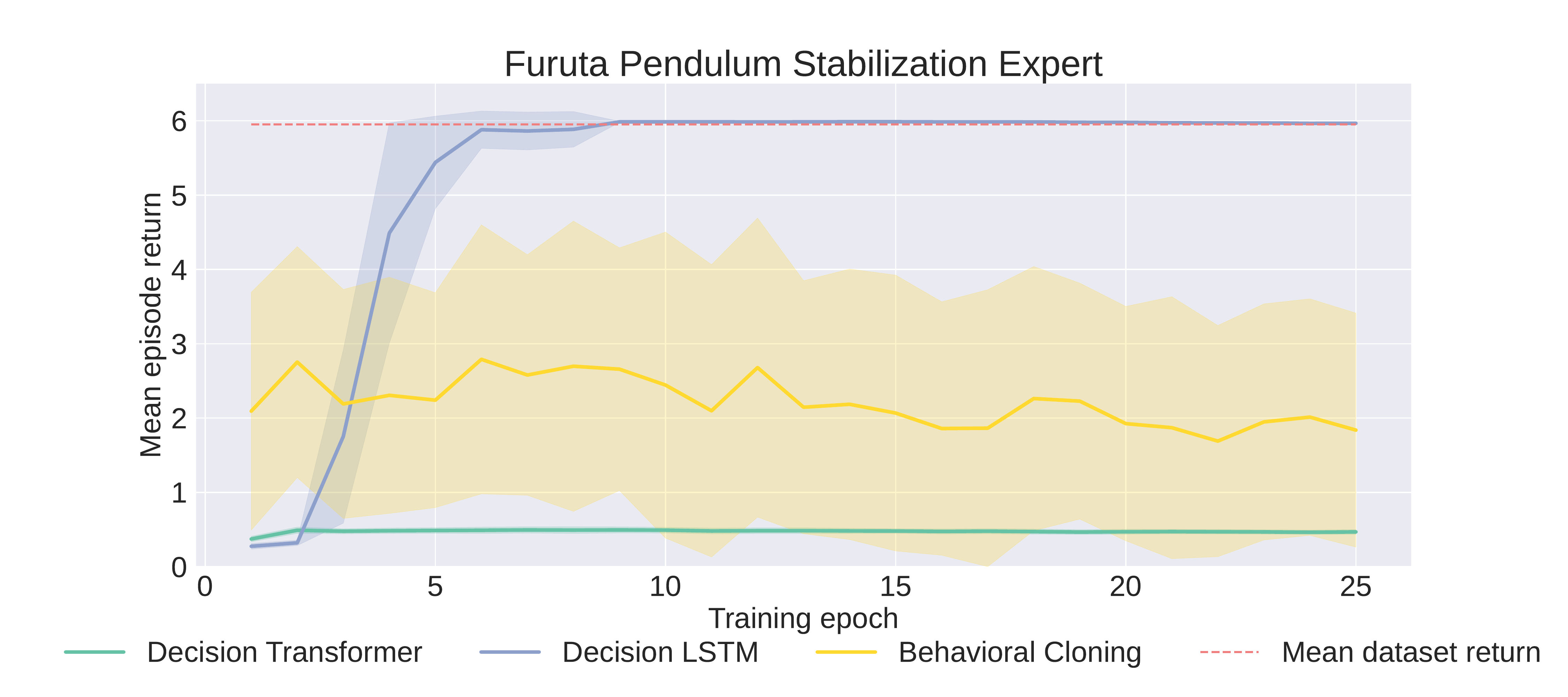}
        \caption[Network2]%
        {{\small Learning curves / expert data, Furuta stabilization}}    
        \label{fig:expert-curve-stab}
    \end{subfigure}
    \hfill
    \begin{subfigure}[b]{0.475\textwidth}
        \centering
        \includegraphics[width=\textwidth]{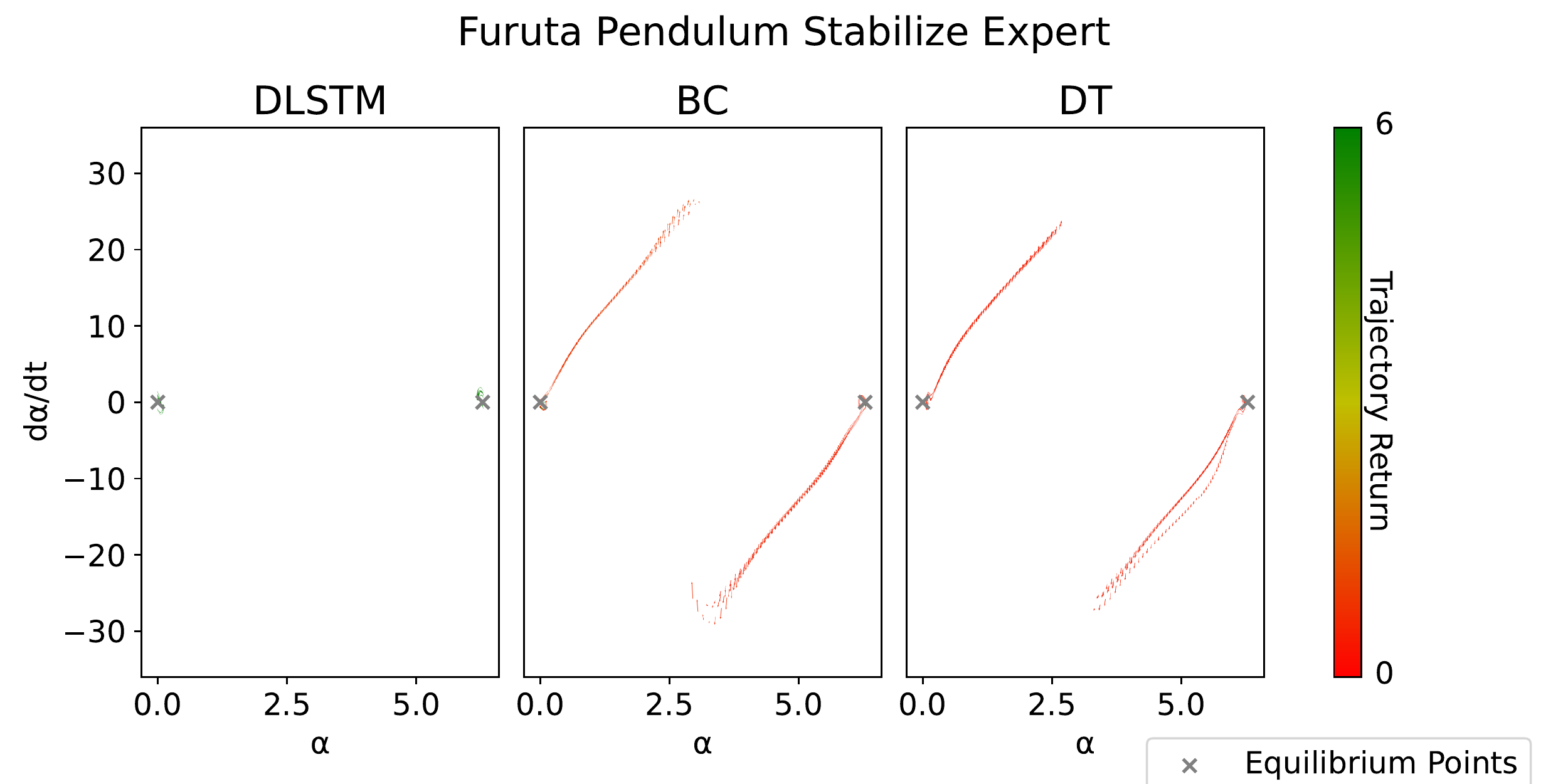}
        \caption[Network2]%
        {{\small Phase portraits / expert data, Furuta stabilization}}    
        \label{fig:expert-phase-stab}
    \end{subfigure}
    \vskip\baselineskip
    \begin{subfigure}[b]{0.475\textwidth}  
        \centering 
        \includegraphics[width=\textwidth]{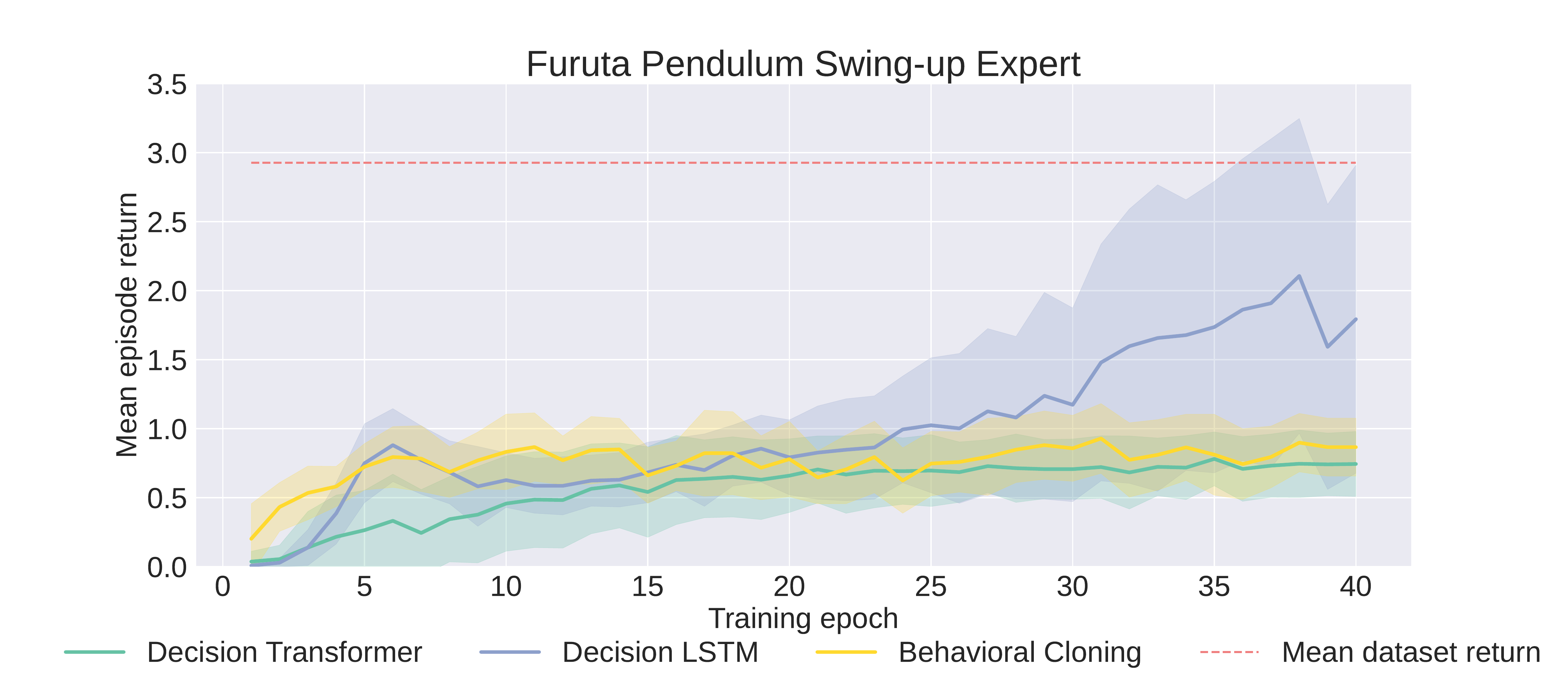}
        \caption[]%
        {{\small Learning curves / expert data, Furuta swing-up}}    
        \label{fig:expert-curve-swup}
    \end{subfigure}
    \hfill
    \begin{subfigure}[b]{0.475\textwidth}
        \centering
        \includegraphics[width=\textwidth]{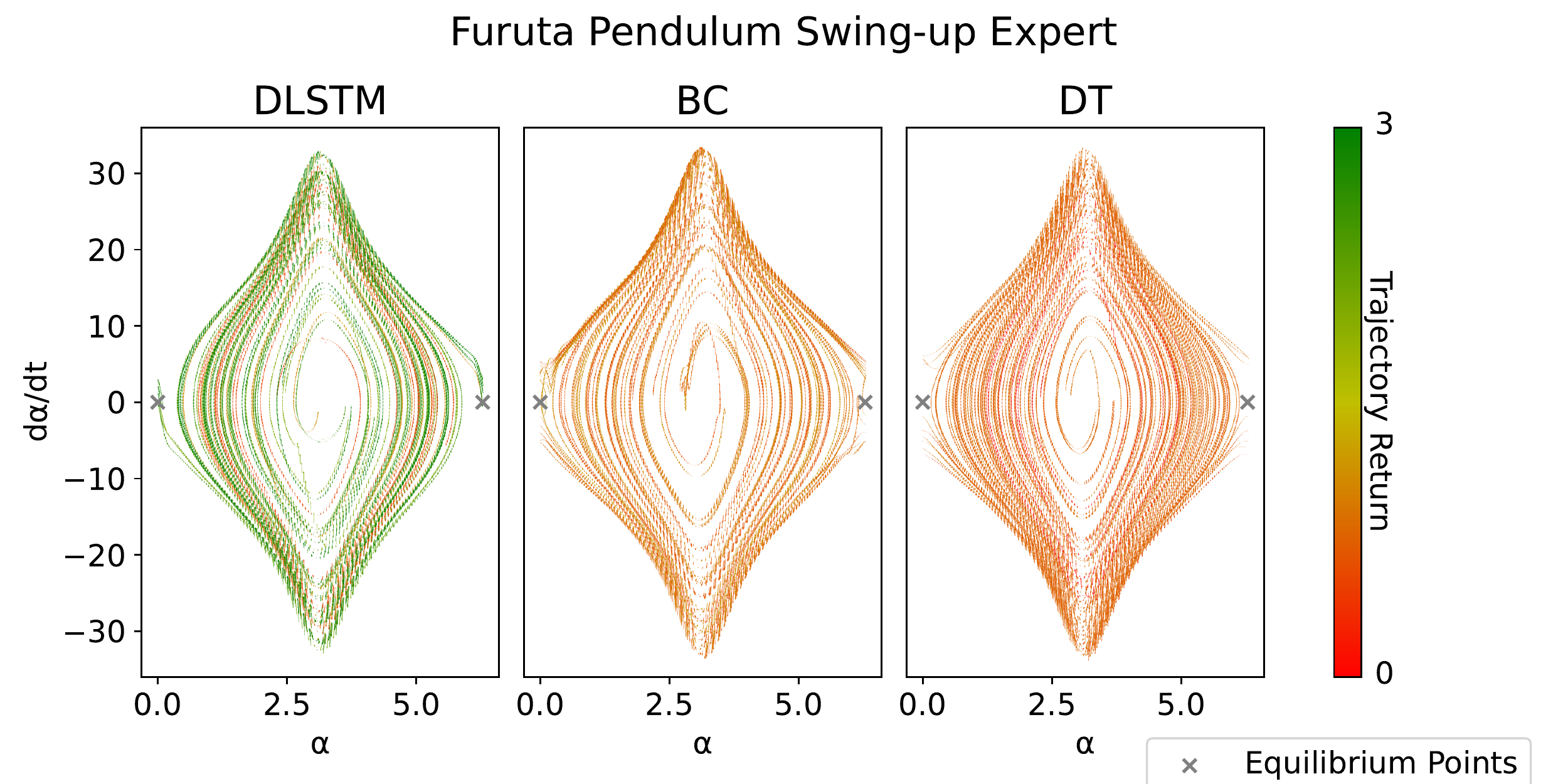}
        \caption[Network2]%
        {{\small Phase portraits / expert data, Furuta swing-up}}    
        \label{fig:expert-phase-swup}
    \end{subfigure}
    \caption[Learning curves on expert data.]
    {Learning curves (left) and phase portraits (right) for DT, DLSTM, and BC trained on \emph{expert} data for stabilization (top) and swing-up (bottom) tasks.
    Especially stabilization (a) appears to pose a challenge to DT and BC, whereas DLSTM quickly achieves the mean dataset return and is able to keep the pendulum at equilibrium, as seen in (b).
    The swing-up task (c) is significantly harder, and again only DLSTM manages to reach sufficiently high return, albeit not in all runs.
    Phase portraits (d) show that DLSTM is the only model which is able to stabilize the pendulum.}
    \label{fig:expert_results}
    \vspace{-1em}
\end{figure}

Table~\ref{tab:experiment-final-results} shows the performance of DT, DLSTM, and BC on simulated swing-up and stabilization tasks.
Decision LSTM outperforms both Decision Transformer and Behavioral Cloning in most experiments, and performs on par with BC on OpenAI pendulum swing-up.
The superior performance of DLSTM is especially apparent in the more challenging Furuta pendulum environment, both on the swing-up and stabilization tasks, on which DT and BC fail to reach the expert performance.

Figures \ref{fig:expert-curve-stab} and \ref{fig:expert-curve-swup} show the learning curves
on stabilization and swing-up tasks in the Furuta pendulum environment.
DLSTM is the only model that manages to solve both tasks, albeit the swing-up is not successful in every run.
Figures~\ref{fig:expert-phase-stab} and \ref{fig:expert-phase-swup} show the phase portraits of the trained models.
In the stabilization task in Figure \ref{fig:expert-phase-stab}, the pendulum starts in the upright unstable equilibrium state, indicated by the points $\alpha = 0$ and $\alpha = 2\pi$.
DLSTM achieves stabilization and high returns in all evaluation episodes.
Meanwhile, BC sometimes achieves stabilization (indicated by green trajectories, high return), but often fails (red trajectories, low return).
DT always fails at the stabilization task: trajectories diverge from the equilibrium state in all episodes.
On the swing-up task (Figure~\ref{fig:expert-phase-swup}), BC and DT manage to bring the pendulum to the upright position, but fail to stabilize it.
DLSTM, on the other hand, is able to swing-up and stabilize the pendulum, albeit not at expert level in every run.

\paragraph{Simulation results on replay data.}

\begin{table}[t]
    \centering
    \caption[Simulation results on replay data.]{Simulation results on \emph{replay} data. DLSTM is the only model which achieves better mean episode return than the average return $\overline{G_{\mathrm{Data}}}$ of the training dataset.}
    {\fontsize{7}{7} \selectfont            
    \begin{tabular}{ccc|ccc} \toprule
    \multicolumn{3}{c}{} & \multicolumn{3}{c}{\textbf{Evaluation Episode Returns}}  \\
    \textbf{Environment} & \textbf{Dataset} & \textbf{$\overline{G_{\mathrm{Data}}}$} &  \textbf{DT} &  \textbf{DLSTM} &  \textbf{BC} \\ \hline 
    OpenAI Pendulum Swing-up & Replay & -837.35 $\pm$ 414.12 & -1083.78 $\pm$ 346.79 & \textbf{-569.89 $\pm$ 568.49} & -815.41 $\pm$ 577.83 \\ 
    Furuta Pendulum Swing-up & Replay & 1.56 $\pm$ 1.70 & 0.51 $\pm$ 0.25 & \textbf{1.30 $\pm$ 1.28} & 0.89 $\pm$ 0.83 \\ \bottomrule
    \end{tabular}
    \label{tab:replay-results}
    }
    \vspace{-1em}
\end{table} 

\begin{figure}[t]
    \centering
    \begin{subfigure}[b]{0.475\textwidth}   
        \centering 
        \includegraphics[width=\textwidth]{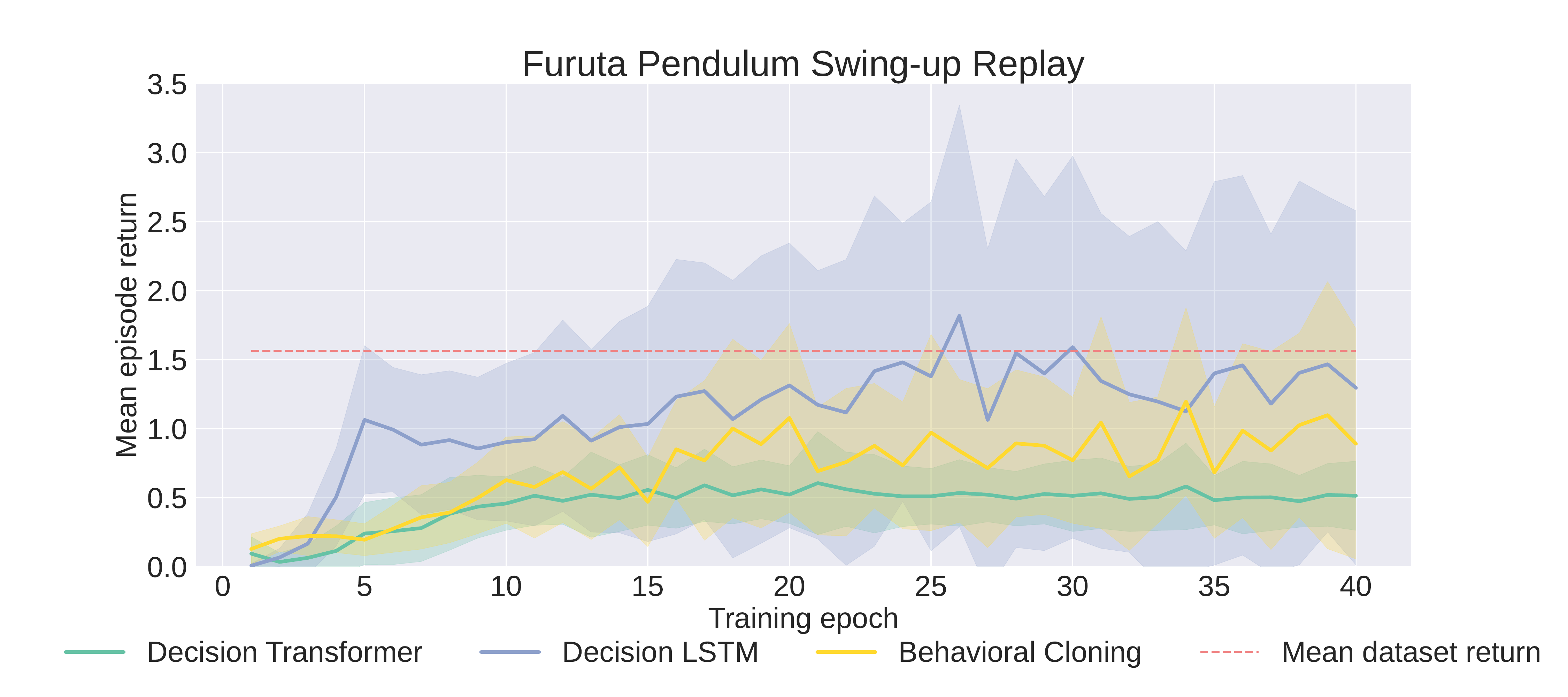}
        \caption[]%
        {{\small Learning curves / replay data, Furuta swing-up}}    
        \label{fig:replay-curve-swup}
    \end{subfigure}
    \hfill
    \begin{subfigure}[b]{0.475\textwidth}
        \centering
        \includegraphics[width=\textwidth]{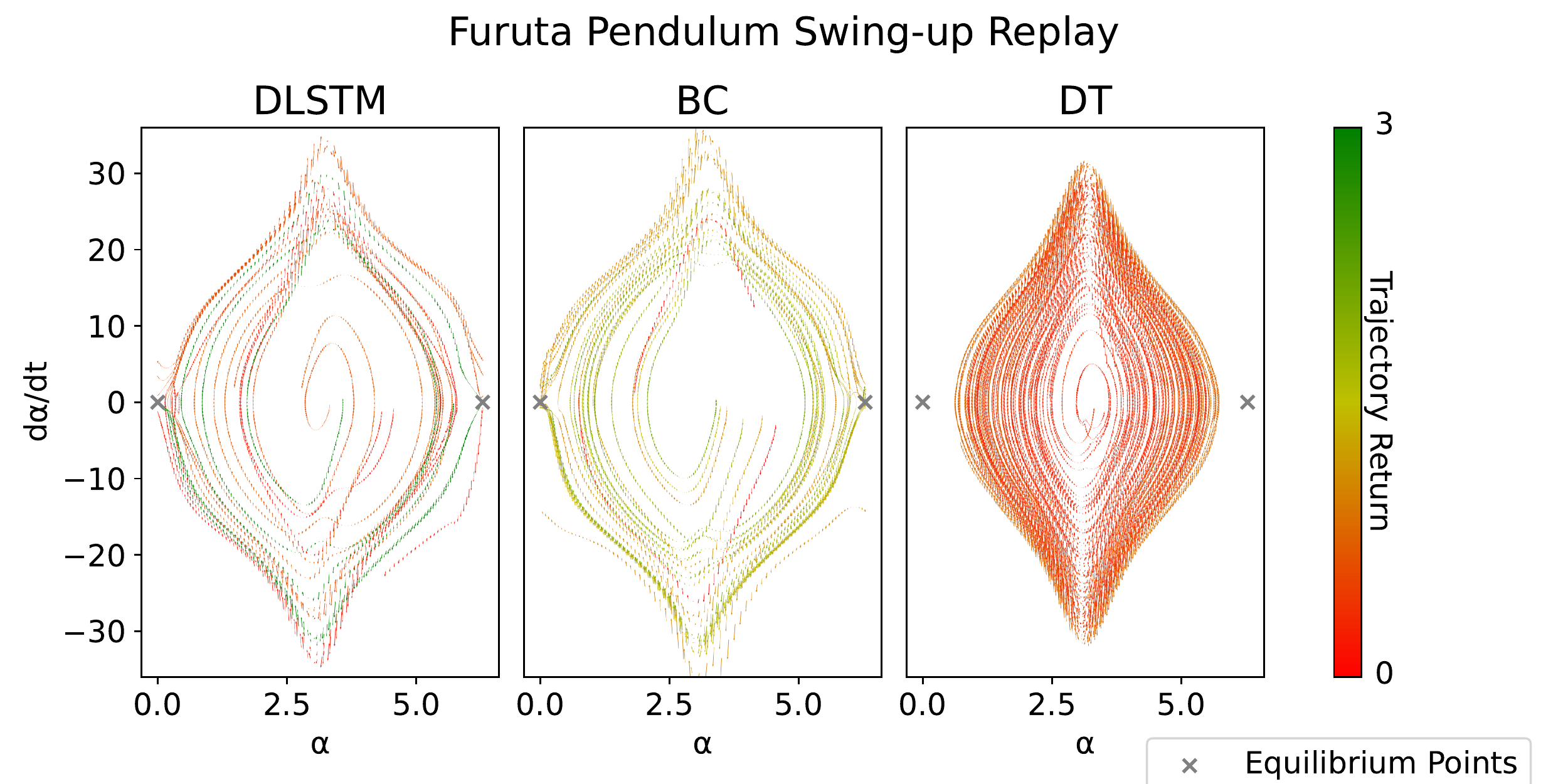}
        \caption[Network2]%
        {{\small Phase portraits / replay data, Furuta swing-up}}    
        \label{fig:replay-phase-swup}
    \end{subfigure}
    \caption[Learning curves on replay data.]
    {Learning curves (left) and phase portraits (right) for DT, DLSTM, and BC trained on \emph{replay} data for the Furuta swing-up task.
    Notably, only DLSTM is able to achieve higher returns than the mean dataset return (the shaded blue area in (a) goes higher than the dotted red line).
    The phase portraits (b) again show that DLSTM is the only model that stabilizes the pendulum.}
    \label{fig:replay_results}
    \vspace{-1em}
\end{figure}

On the replay data, DLSTM again achieves better performance than the other models (Table~\ref{tab:replay-results}).
Notably, DLSTM is the only model which is able to improve upon the demonstrations, i.e., achieve a higher return than in the  training dataset.
Figure~\ref{fig:replay-curve-swup} shows the corresponding learning curves and Figure~\ref{fig:replay-phase-swup} the phase portraits.
The results are similar to Figure~\ref{fig:expert-phase-swup}.

\vspace{-0.5em}
\paragraph{Real platform results.}

\begin{table}[t]
    \centering
    \caption[Experimental results on Furuta pendulum real robot (FPRR).]{Experimental results on the real Furuta pendulum (denoted FPRR) on expert data.
    DLSTM matches expert performance on the stabilization task and achieves high but less than expert reward on swing-up and swing-up with PD-stabilization tasks.
    Only DLSTM was evaluated with PD-stabilization, because DT and BC failed to swing-up the real Furuta pendulum.
    } 
    {\fontsize{7}{7} \selectfont        
    \begin{tabular}{ccc|ccc} \toprule
    \multicolumn{3}{c}{} & \multicolumn{3}{c}{\textbf{Evaluation Episode Returns}}  \\
    \textbf{Environment} & \textbf{Dataset} & \textbf{$\overline{G_{\mathrm{Data}}}$} &  \textbf{DT} &  \textbf{DLSTM} &  \textbf{BC} \\ \hline 
    FPRR Swing-up & Expert & 2.93 $\pm$ 0.63 & 0.38 $\pm$ 0.15 & \textbf{1.11 $\pm$ 0.52} & 0.22 $\pm$ 0.18 \\
    FPRR Swing-up with PD stabilization & Expert & 2.93 $\pm$ 0.63 & $-$ & \textbf{2.17 $\pm$ 0.60} & $-$ \\
    FPRR Stabilization & Expert & 5.95 & 0.38 $\pm$ 0.08 & \textbf{5.98 $\pm$ 0.00} & \textbf{5.96 $\pm$ 0.02} \\ \bottomrule
    \end{tabular}
    \label{tab:experiment-rr-final-results}}
    \vspace{-1em}
\end{table} 

\begin{figure}[t]
    \centering
    \begin{subfigure}[b]{0.475\textwidth}   
        \centering 
        \includegraphics[width=\textwidth]{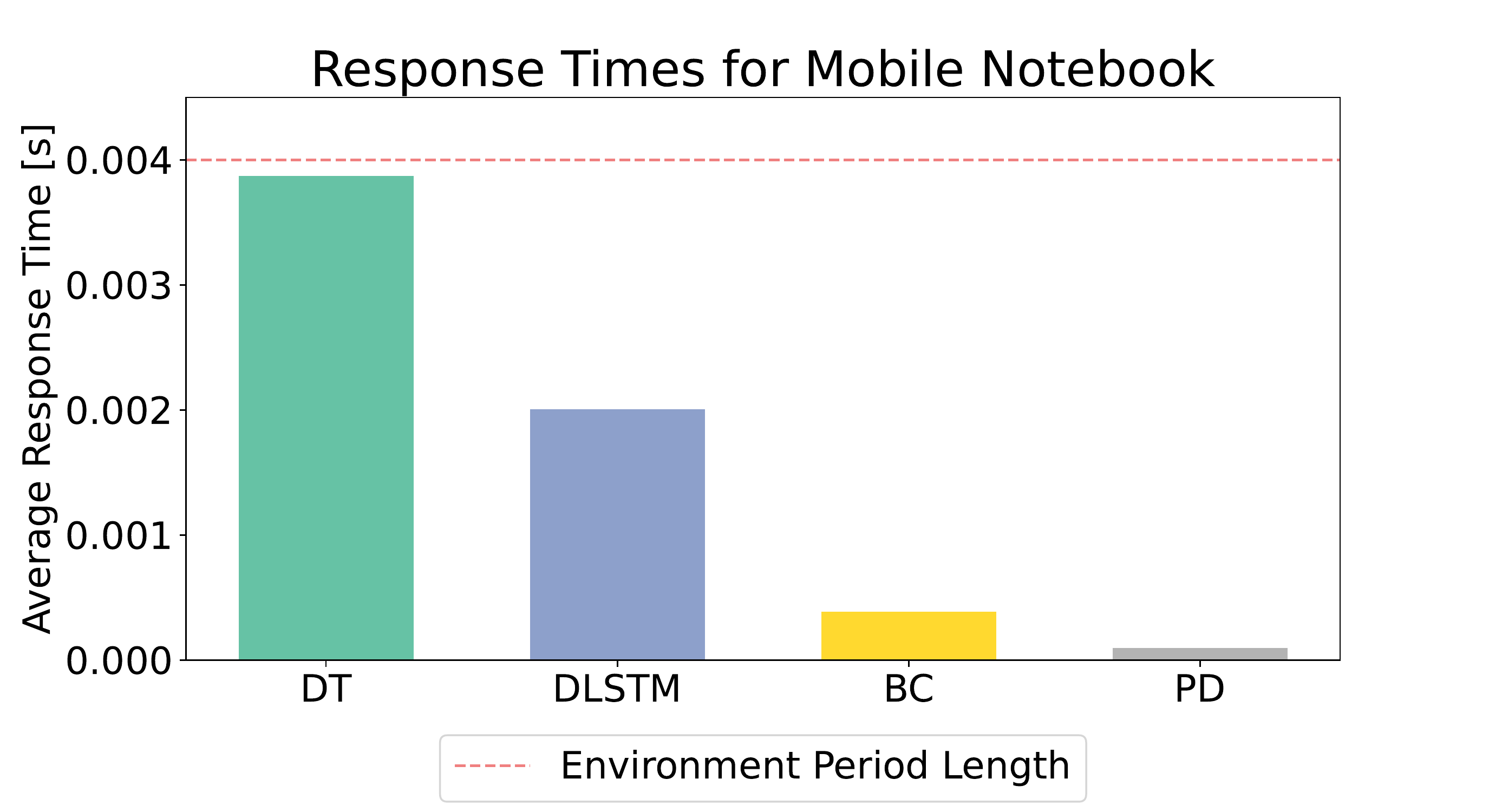}
        \caption[]%
        {{\small Inference times on laptop with Intel(R) Core(TM) i5-7200U CPU, 2 cores @ 2.50 GH.}}
        \label{fig:inference-mobile-pc}
    \end{subfigure}
    \hfill
    \begin{subfigure}[b]{0.475\textwidth}
        \centering
        \includegraphics[width=\textwidth]{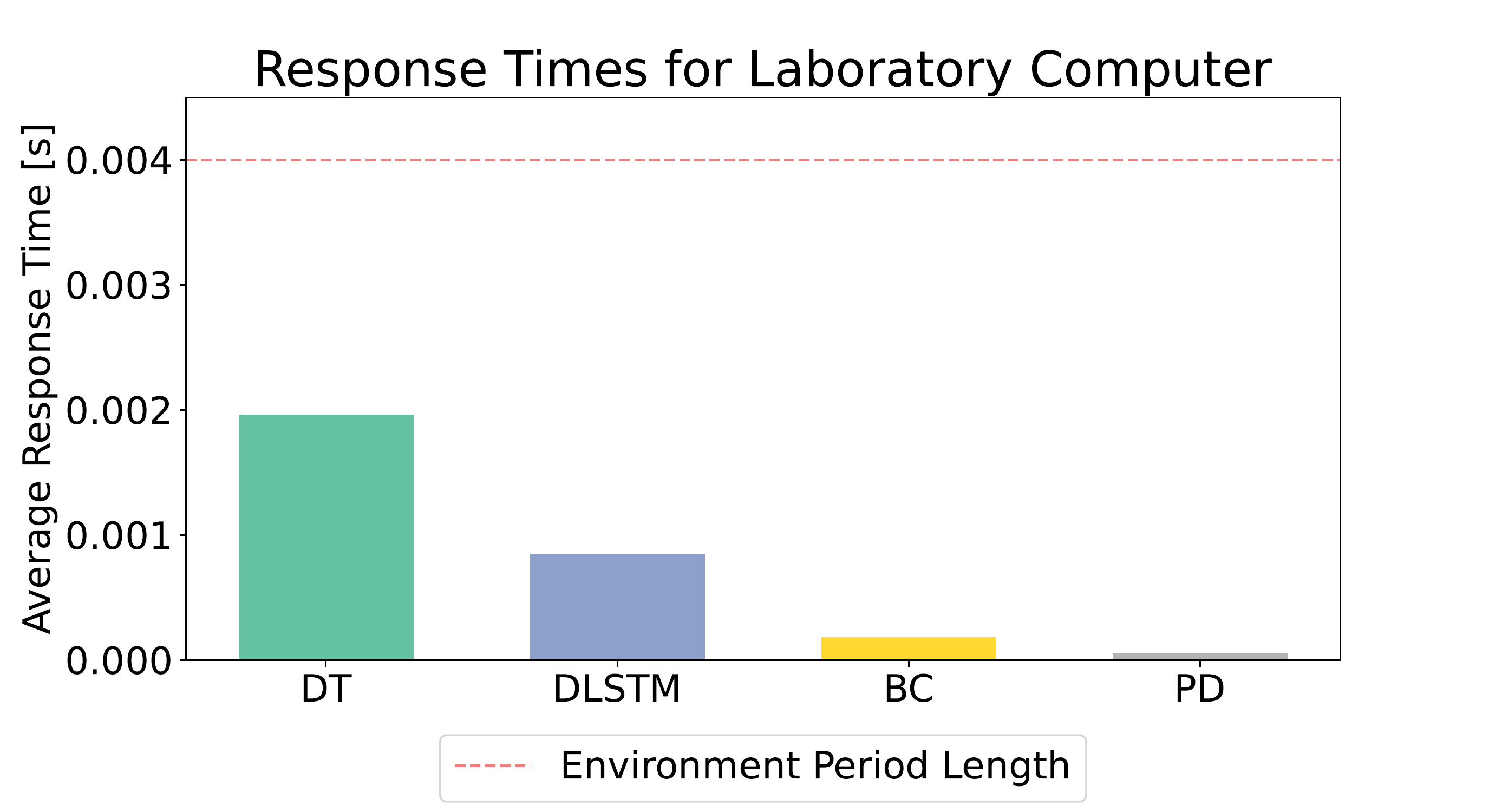}
        \caption[]%
        {{\small Inference times on computer with Intel(R) Core(TM) i7-9700K CPU, 8 cores @ 3.60 GH.}}    
        \label{fig:inference-lab-pc}
    \end{subfigure}
    \caption[Mean inference times.]
    {Mean inference times for DT, DLSTM, BC, and a PD controller on a real Furuta pendulum
    commanded from a laptop (a) and a stationary PC (b).
    In both cases, the mean inference time is below the control interval $0.004$s for all methods.
    However, DT takes twice the time compared to DLSTM.}
    \label{fig:response_times}
    \vspace{-1em}
\end{figure}

For a real-world evaluation, the models trained on an expert dataset recorded on the real Furuta pendulum platform are evaluated in this environment.
Table~\ref{tab:experiment-rr-final-results} indicates that DLSTM significantly outperforms DT and BC.
On the standard swing-up task, DLSTM achieves stabilization and high return in many but not all episodes, while the other models fail to bring the pendulum to the  upright position altogether.
Despite successful swing-up, DLSTM fails at stabilizing the pendulum in most cases, therefore the episode return is lower than in the expert data.
Such performance gap can be explained by the sim-to-real discrepancy and the real-time requirements of the physical platform.

To show that DLSTM is both capable of swinging up and stabilizing the pendulum, we test its capabilities on the respective tasks independently.
First, we let the DLSTM policy swing up the pendulum and, after a certain pose is reached, let a PD controller take over and stabilize the pendulum.
The second row in Table \ref{tab:experiment-rr-final-results} shows that this combination of the controllers leads to higher mean returns than before.
Apparently, DLSTM has problems dealing with the high velocities which occur during the swing-up phase but which were not observed in the simulation training data.

Finally, we evaluate all models on the pure stabilization task on the real Furuta pendulum platform.
We use a given expert policy to swing up and stabilize the pendulum for a short period of time, and then let the respective models (DT, DLSTM or BC) take over to continue stabilizing the pendulum.
DLSTM and BC achieve expert performance, while DT fails on this task ($3$rd row in Table~\ref{tab:experiment-rr-final-results}).

\vspace{-0.5em}
\paragraph{Real-time capabilities.}
One reason for the worse performance of the considered models on the physical platform compared to simulation may be the time delays in the real-time control loop.
To investigate this issue,
we compare the mean inference times, i.e., the time needed to generate an action, for the different models in the Furuta pendulum environment.
Figure~\ref{fig:response_times} shows response times measured on a laptop and a stationary PC.
The horizontal orange line shows the maximum allowed time for action generation in the real-time loop at $250$Hz.
This control frequency is necessary to enable stable and reliable pendulum stabilization.
DT and DLSTM have higher inference times compared to BC and a PD-controller.
DT on average takes twice the time compared to DLSTM.

\paragraph{Influence of return-to-go values.}

\begin{figure}[ht]
\vspace{-1em}
    \centering
    \begin{subfigure}[b]{0.4\textwidth}
        \centering
        \includegraphics[width=\textwidth]{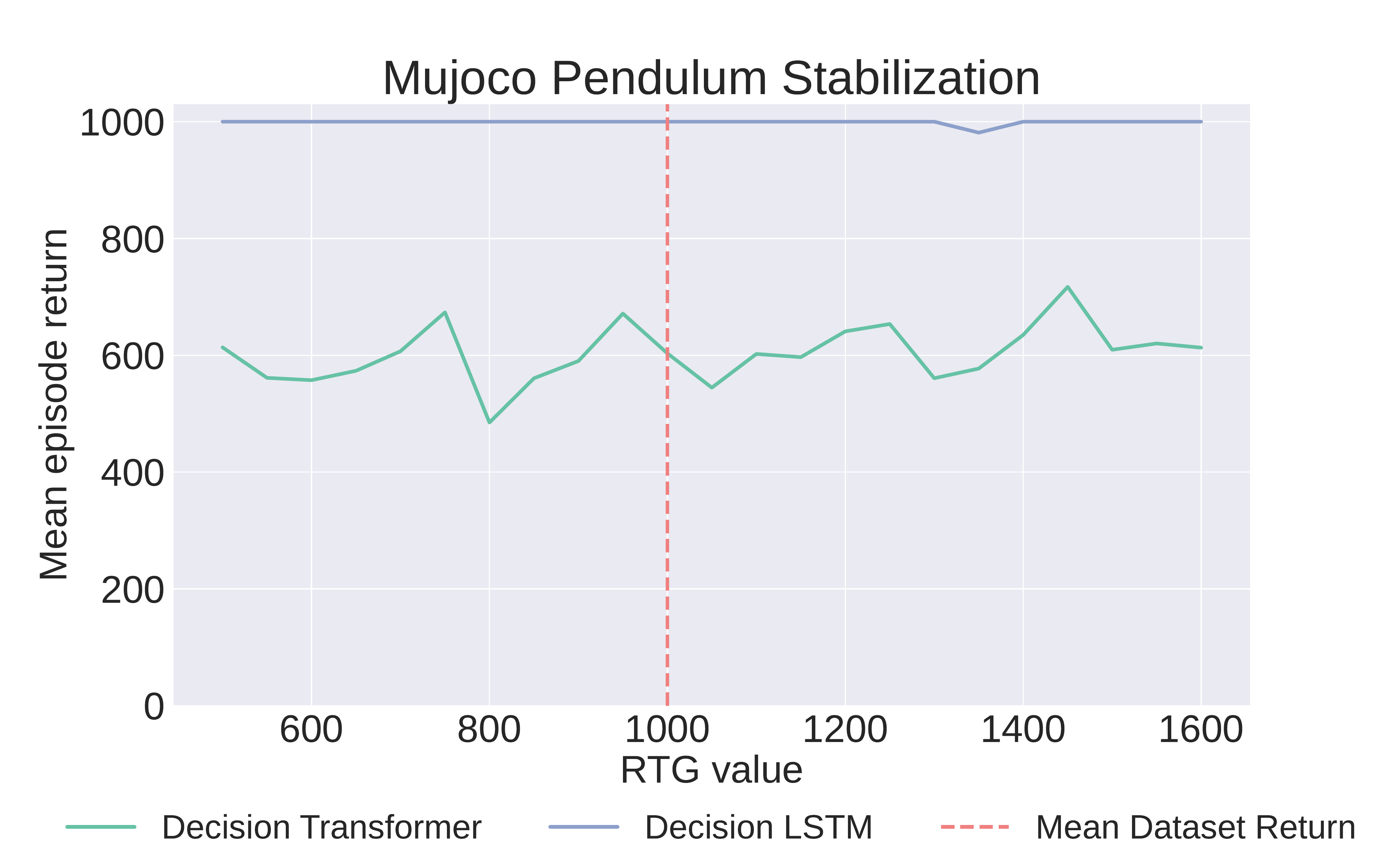}
        \caption[Network2]%
        {{\small RTG influence in Mujoco InvPend}}    
        \label{fig:mean and std of net14}
    \end{subfigure}
    \begin{subfigure}[b]{0.4\textwidth}  
        \centering 
        \includegraphics[width=\textwidth]{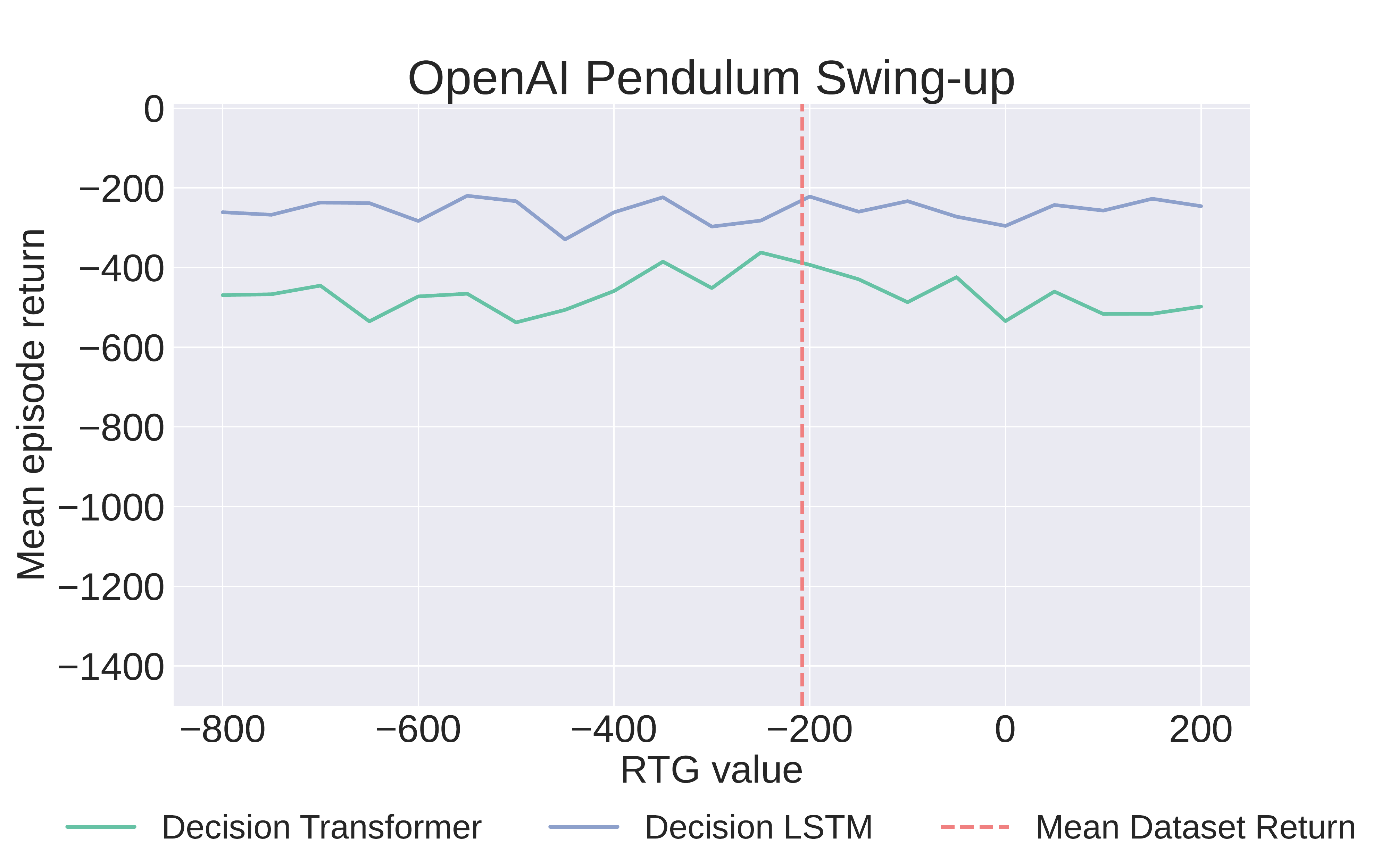}
        \caption[]%
        {{\small RTG influence in OpenAI pendulum}}    
        \label{fig:mean and std of net24}
    \end{subfigure}
    \vskip\baselineskip
    \begin{subfigure}[b]{0.4\textwidth}   
        \centering 
        \includegraphics[width=\textwidth]{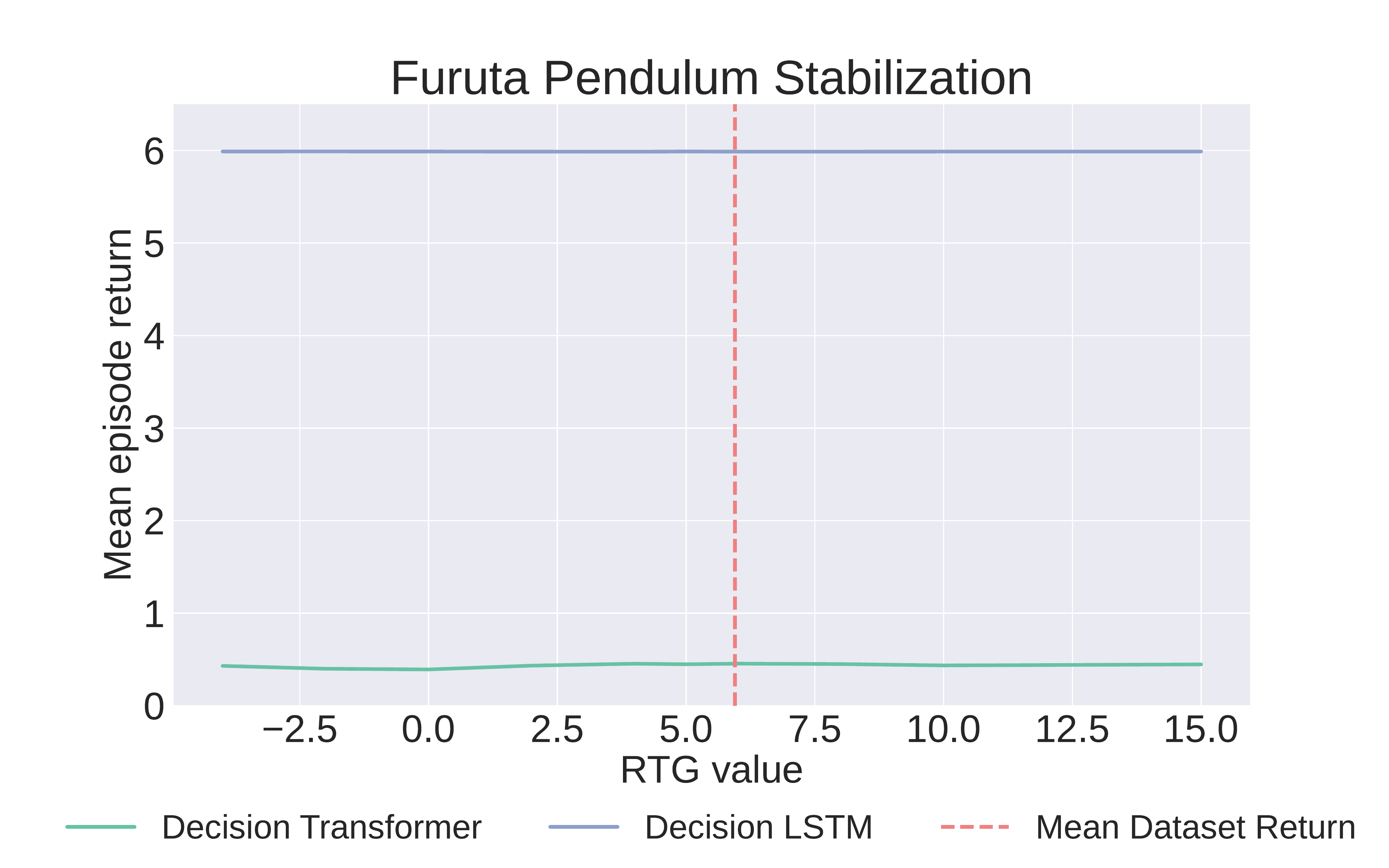}
        \caption[]%
        {{\small RTG influence in Furuta stabilization}}    
        \label{fig:mean and std of net34}
    \end{subfigure}
    \begin{subfigure}[b]{0.4\textwidth}   
        \centering 
        \includegraphics[width=\textwidth]{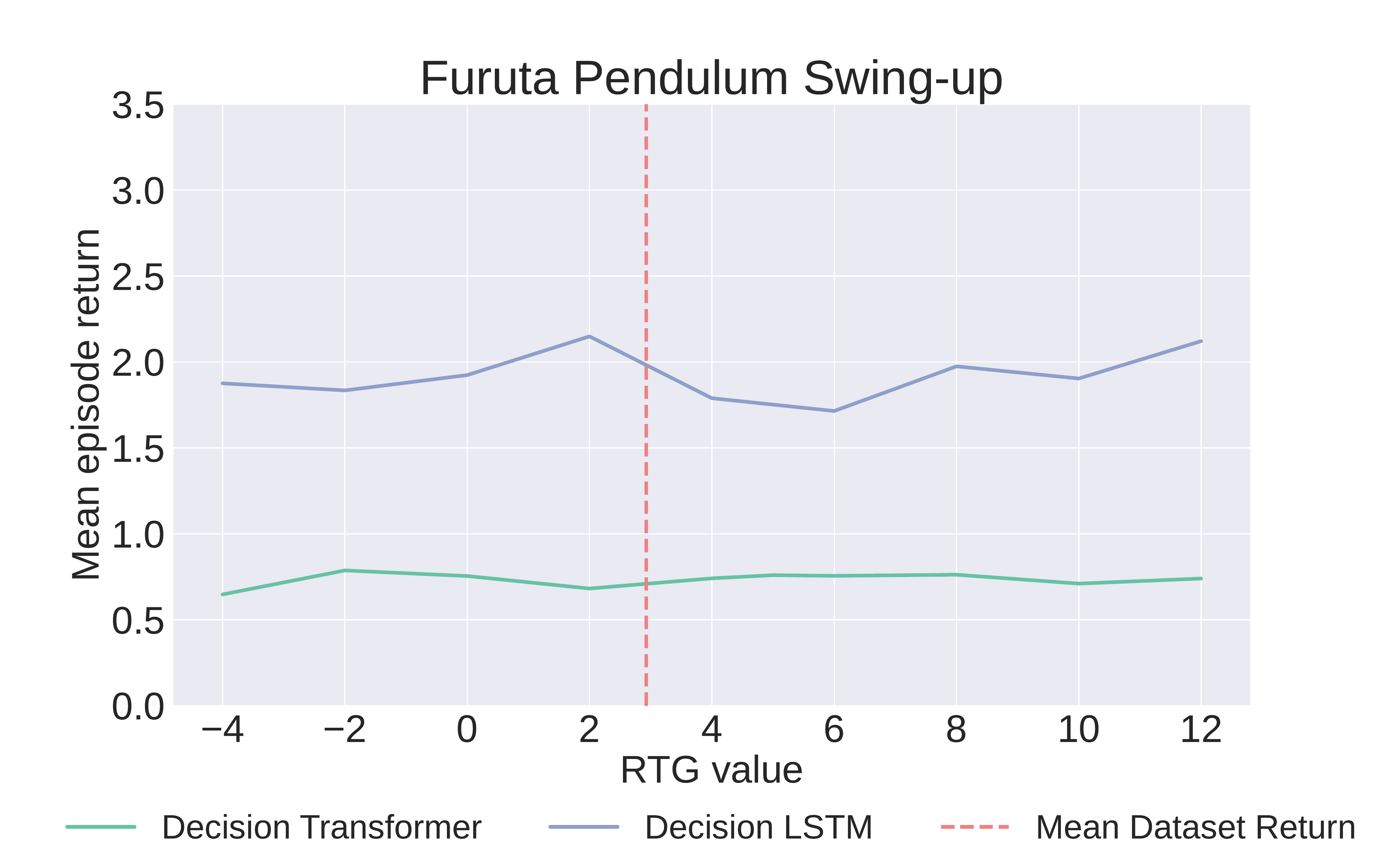}
        \caption[]%
        {{\small RTG influence in Furuta swing-up}}    
        \label{fig:mean and std of net44}
    \end{subfigure}
    \caption[RTG effects]
    {Effects of the desired return-to-go (RTG) values used for conditioning action generation by DT and DLSTM on the mean episode return in different environments.
    The query RTG value appears to have no influence on the episode return, which indicates that the models are not making use of the RTG value for action generation.} 
    \label{fig:results_rtg_comp}
    \vspace{-1em}
\end{figure}

An important feature of DT is the use of return-to-go (RTG) values, i.e., one can in principle control the optimality of the generated trajectories:
at evaluation time, the RTG value specifies the desired expected return, and therefore DT should generate trajectories that achieve this desired RTG value.
We perform evaluations to verify whether the RTG value indeed has an influence on the episode return.
Contrary to the findings in~\cite{chen2021decision}, Figure~\ref{fig:results_rtg_comp} indicates no influence of the desired RTG values on the actual returns both for DT and DLSTM.
These results raise the question whether the RTG values are even necessary in the DT architecture and what role they  play.

\section{Related Work}

The original Decision Transformer~\cite{chen2021decision} presents a basic approach to framing RL as sequence modeling problem, allowing for extensions in multiple directions.
In \cite{furuta2022generalized}, a \emph{Generalized Decision Transformer} is introduced, which performs hindsight information matching by generating trajectories that match any statistics of the future trajectories and not only return-to-go values.
The \emph{Online DT}~\cite{zheng2022online} combines offline pre-trained DT models with an online fine-tuning procedure, thereby overcoming the distributional shift inherent in offline RL and affecting the DT.
A similar online DT approach but in a \emph{multi-agent} setting is proposed in~\cite{meng2021offline}.
\emph{Transfer learning} for the DT architecture is addressed in~\cite{reid2022wikipedia}, where the DT is pre-trained on data from other domains and modalities, e.g., Wikipedia articles, and subsequently fine-tuned on a given offline RL problem.
On a massive scale, such cross-modal generalization capability of DT was demonstrated in 
\emph{Gato}~\cite{reed2022generalist}.
These results indicate the existence of an underlying universal structure across sequence modeling problems that enables cross-domain and multi-modal transfer learning.

\section{Conclusion}
Our empirical evaluations of Decision Transformer on continuous control tasks such as pendulum swing-up and stabilization have shown that DT struggles on problems that require fine-tuned actions.
The model trained on offline data fails in the online setting and is not able to solve the task on a real system.
On the other hand, the proposed modification of DT, which we call Decision LSTM---and which only differs from DT in that the Transformer is replaced by an LSTM---has shown strong performance in the same environments.
Therefore, we conclude that the advantages of DT observed in prior works may be rather due to the sequence modeling approach than to the particular choice of the prediction module.

The paper does not consider discrete actions (e.g., discretizing continuous actions into bins, or discrete-action domains such as Atari), where transformer-style architectures may have an advantage.
Moreover, performance of DT-like policy architectures can depend on many factors, including the domain/task (transition dynamics and reward function), action parameterization, discretization of RTG, network architecture, etc.
In general, our results only apply to the continuous control tasks and a further investigation is necessary to evaluate our hypothesis on a broader set of domains.

Despite the good performance of DLSTM, it remains an open question whether approaches that frame RL as a sequence modeling problem provide significant advantages over standard Behavioral Cloning.
Our results indicate no correlation between the return-to-go values and the model performance in the stabilization experiments.
Therefore, the effectiveness of RTG values as task-defining inputs that provide hindsight information to the decision architectures in continuous control tasks is unclear.

Finally, to make Decision Transformer and Decision LSTM applicable in real-world settings, the sim-to-real gap and the inference times of the models are crucial.
The inference times of the decision architectures are significantly longer compared to standard BC, which yielded problems in our real-time experiments, making it necessary to investigate the inference times of the models further.
Purely relying on successful simulation runs where the actual inference times of the models are ignored may be a potential cause of problems under real-world conditions.
In settings where Decision Transformer generates actions too slow for the real-time requirements, the proposed Decision LSTM architecture may be preferred due to the faster run time.

\begin{ack}
This project has received funding from BMWSB ZukunftBau under grant Nr. 10.08.18.7-21.34.
Calculations for this research were partially conducted on the Lichtenberg high performance computer of the TU Darmstadt.
\end{ack}

\bibliographystyle{abbrv}
\bibliography{lit}

\appendix
\section{Appendix}

\subsection{Datasets}

Table~\ref{tab:dataset-overview} provides an overview of the datasets used in our experiments.
The behavior policy is given by a PPO agent~\cite{schulman2017proximal}, i.e., the \textit{replay} datasets are comprised of the experiences collected during all epochs of the PPO training, while the \textit{expert} datasets contain demonstrations of the PPO policy in the final epoch, i.e., after it has been trained to expert-level.
As a measure of the quality of demonstrations in the dataset, the expected trajectory return (TQ) values~\cite{Schweighofer2021UnderstandingTE} are used, which quantify the relation between the average return of a trajectory to the maximal return in the dataset.
A high TQ value indicates high quality (i.e., high mean returns) in the dataset, while a low TQ value indicates low-return demonstrations.

\begin{table}[!htb]
    \centering
    {\fontsize{8}{8} \selectfont
    \begin{tabular}{ccccccc}
    \toprule
        \textbf{Name}                    & \textbf{Environment}                   & \textbf{Num. Traj}   & \textbf{Beh. Policy}     & \textbf{Task}     & \textbf{TQ}                 \\ \hline
        mujoco-inverted-pendulum-expert     & Mujoco Inv. Pendulum           & 500           & PPO      &  Stabilization  & 1.00    \\    \hline
        openai-pendulum-expert     & OpenAI Pendulum           & 250           & PPO      &  Swing up  & 0.83   \\    
        openai-pendulum-replay     & OpenAI Pendulum           & 100           & PPO      &  Swing up  & 0.32    \\    \hline
        furuta-pendulum-stabilize-expert     & Furuta Pendulum           & 500           & PPO      &  Stabilization  & 0.99    \\    
        furuta-pendulum-swing-up-expert     & Furuta Pendulum           & 500           & PPO      &  Swing up  & 0.49    \\    
        furuta-pendulum-swing-up-replay     & Furuta Pendulum           & 515           & PPO      &  Swing up  & 0.29   \\    \bottomrule
       
    \end{tabular}}
    \caption[Overview of the used training datasets for the stabilization experiments.]{Overview of the used training datasets for the stabilization experiments.
    }
    \label{tab:dataset-overview}
\end{table}

\subsection{Hyperparameter Settings}

For the experiments, the default hyperparameter settings from \cite{chen2021decision} were used, as shown in Table \ref{tab:experiment-hyperparams}.

\begin{table}[!htb]
    \centering
    {\fontsize{10}{10} \selectfont
    \begin{tabular}{cccc}
    \toprule
         & \textbf{DT} & \textbf{DLSTM} & \textbf{BC} \\ \midrule
        Context length $K$ & 20 & 20 & 20 \\
        Number of hidden layers & 3 & 3 & 3\\
        Hidden layer size & 128 & 128 & 256\\
        Batch size & 64 & 64 & 128\\
        Number of training steps per training epoch & 3000 & 3000 & 3000 \\
        Input normalization & yes & yes & yes\\
        Dropout & 0.1 & 0.1 & 0\\
        Activation function & $\tanh$ & $\tanh$ & $\tanh$\\
        Learning rate & \num{3e-5} & \num{3e-5} & \num{3e-5}\\
        Number of attention heads & 1 & - & -\\ \bottomrule
    \end{tabular}
    \caption[Overview of the used hyperparameters for the different evaluated architectures.]{Overview of the used hyperparameters for the different evaluated architectures.} 
    \label{tab:experiment-hyperparams} }
\end{table}

\label{chap:app}

\end{document}